\documentclass{article}

\usepackage{PRIMEarxiv}

\usepackage[utf8]{inputenc} 
\usepackage[T1]{fontenc}    
\usepackage{hyperref}       
\usepackage{url}            
\usepackage{booktabs}       
\usepackage{amsfonts}       
\usepackage{nicefrac}       
\usepackage{microtype}      
\usepackage{lipsum}
\usepackage{fancyhdr}       
\usepackage{graphicx}       
\graphicspath{{media/}}     

\title{Analysing Ultra-Wide Band Positioning for Geofencing in a Safety Assurance Context
}

\author{
  \href{https://orcid.org/0000-0002-2469-0224}{\includegraphics[scale=0.06]{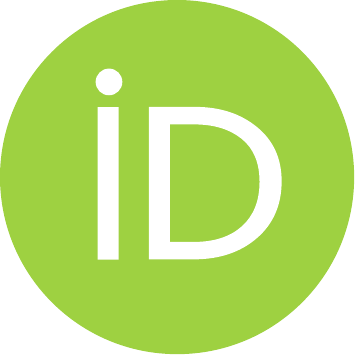}\hspace{0.5mm}Victoria J. Hodge}\thanks{Corresponding author: Dr Victoria Hodge, Department of Computer Science,
        University of York, Heslington, York, UK. \textit{email: victoria.hodge@york.ac.uk}}, \href{https://orcid.org/0000-0001-7347-3413}{\includegraphics[scale=0.06]{orcid.pdf}\hspace{0.5mm}Richard Hawkins}, \hspace{1 mm}James Hilder and \href{https://orcid.org/0000-0003-2736-8238}{\includegraphics[scale=0.06]{orcid.pdf}\hspace{0.5mm}Ibrahim Habli} \\
  Department of Computer Science \\
  University of York \\
  York, UK\\
  \texttt{\{victoria.hodge, richard.hawkins, james.hilder,  ibrahim.habli\}@york.ac.uk} \\

}

\begin{document}
\maketitle

\begin{abstract}
There is a desire to move towards more flexible and automated factories. To enable this, we need to assure the safety of these dynamic factories. This safety assurance must be achieved in a manner that does not unnecessarily constrain the systems and thus negate the benefits of flexibility and automation. We previously developed a modular safety assurance approach, using safety contracts, as a way to achieve this. In this case study we show how this approach can be applied to Autonomous Guided Vehicles (AGV) operating as part of a dynamic factory and why it is necessary. We empirically evaluate commercial, indoor fog/edge localisation technology to provide geofencing for hazardous areas in a laboratory. The experiments determine how factors such as AGV speeds, tag transmission timings, control software and AGV capabilities affect the ability of the AGV to stop outside the hazardous areas. We describe how this approach could be used to create a safety case for the AGV operation.
\end{abstract}


\section{Introduction}

As manufacturing becomes increasingly automated and requires increased flexibility, there is a move to more dynamic factories. This necessitates dynamic supply chain solutions where factories manufacture a broad range of products using frequent reconfiguration of the factory layout to optimise production, efficiency and costs. This dynamism brings particular challenges when attempting to provide assurance for the safety of operation prior to deployment.

Dynamic factory operations require increased digitisation to provide increased flexibility and automation. The implementation of this often requires extensive use of commercial off the shelf (COTS) systems and components, which provide accessible and competitively priced capabilities, but often lack the rigorous certification of components developed specifically for safety-related tasks \cite{jaradat2017challenges}. Dynamic factories are also inherently interconnected, adding to the complexity of the safety analysis and assurance task. It is crucial that these safety challenges are addressed, particularly for autonomous applications \cite{burton2020mind}. It is only if the safety of dynamic factories can be assured that they can be fully embraced. It is also vital that safety assurance is achieved in a manner that does not unnecessarily constrain the systems and thus negate the benefits of flexibility and automation. 

This paper analyses our approach to safety assurance for dynamic factories using a case study. We aim to achieve a balance between safety assurance and viability (cost and efficiency). For this. we consider factory operations that use autonomous guided vehicles (AGV) to move products and components around in an efficient manner. Unlike AGVs in traditional factories, the AGV operation is not constrained to fixed routes, and cannot rely on a fixed infrastructure or environment. In addition AGVs will often have to operate in collaborating teams and in cooperation with humans. To assure factory safety, it must be proven that AGVs can navigate safely by not colliding with other objects or humans present in the factory and by keeping away from potentially changing hazardous areas (danger zones). Geofencing is a commonly used location-based approach which we analyse to control the risk in hazardous areas of a factory. Factories are indoor environments so geofencing requires indoor localisation capability such as through the use of Ultra-Wide Band (UWB) IoT technology. Shule et al. \cite{shule2020uwb} state that ``\emph{UWB has the potential to become a standard technology for relative positioning and ranging in multi-robot systems, having been applied to a
wide variety of scenarios in recent years}''. 

We undertake an experimental evaluation in a laboratory setting using AGVs that must not encroach into a danger zone. Our platform uses commercial hardware for UWB geofencing and an AGV built from COTS hardware plus PC software we developed to analyse the data and communicate with the AGV. Commercial AGVs are expensive so using COTS hardware for the AGV is more economically viable if AGVs are to be used more widely. It also allows us extensive control and enables a more thorough analyses of a number of aspects of AGVs and how they affect safety constraints.

Our main contributions are: we run a lab-based case study of AGV operations making use of commercial IoT geofencing technology and relate our experimental results to the specification of safety contracts. From these evaluations, we consider a modular safety assurance approach for dynamic factories.

The rest of the paper is organised as follows: section \ref{sec:safetyAssurance} provides background for our overall approach that we adopt to safety assurance of dynamic factories. We introduce a scenario that is the subject of our experimental study and describe the safety requirements that we derived in section \ref{sec:scenario}. Section \ref{sec:evaluating} describes our experimental design and evaluates the results. Section \ref{sec:related} considers related indoor localisation work and section \ref{sec:future} details future work.

\section{Safety Assurance of Dynamic Factories}\label{sec:safetyAssurance}


It is crucial that the safety risk associated with factory operations is sufficiently analysed, controlled and monitored \cite{jaradat2017challenges}. Many of the same characteristics that make dynamic factories desirable from a technical perspective also pose significant safety assurance challenges. Dynamic factories are designed to be flexible and easily reconfigured. This makes safety analysis, which is traditionally conducted prior to operation, a challenge since the state of the system is hard to predict in advance \cite{javed2020towards}. Control in dynamic factories is also often widely distributed, with limited centralised control. This can make the task of determining and enforcing safety requirements more challenging. Compounding this, factory operators will often have little control over the design and evolution of the commercial systems and components that are used. This can significantly weaken safety assurance due to a high degree of uncertainty about the actual performance or behaviour of these commercial components. 

Our approach to safety assurance is modular and based on the use of safety contracts \cite{fenn2007safety} between elements of the factory.  The German automation technology supplier ‘PILZ’ \cite{pilz} places emphasis on the necessary modular certification of the individual factory devices (PILZ uses the term Safety 4.0 to indicate modular safety solutions). Most of the literature however focuses on  dependability in general but without much focus on safety which is challenging in dynamic environments \cite{javed2020towards}.

\subsection{Safety Cases for Dynamic Factories}

 In many safety-related domains, such as railways or nuclear, it is common practice to develop an explicit safety case for systems. The safety case sets out the argument as to why a system is considered safe to operate in a particular environment, and provides evidence to support the argument (such as results from analysis or testing) \cite{kelly2004goal}. Safety cases are particularly used for novel systems where there is limited established best practice for assuring safety, where the explanation for why the system is safe must be clearly communicated \cite{sujan2016should}. Given the novelty of dynamic factories, we propose that an explicit safety case should be provided.
 
Making factories more dynamic often requires a shift from standalone systems to edge or fog networks of devices and services performing, cooperatively, a number of functionalities \cite{kagermann13}. Creation of a safety case will therefore generally need to be cooperative in the sense that a safety case for a dynamic factory cannot be built by a single stakeholder or organisation.  The supplier organisations, (e.g. the supplier of a smart sensor used in the factory), have the best knowledge of the properties and characteristics of their components and should define a set of safety guarantees for different usages. However, there is a limit to what the supplier organisation is able to provide out-of-context from the particular factory or operation; remember that such components are often general-purpose commercial products that may be used in many different applications.  The suppliers can provide assurance for the component (e.g. through the use of UL/ CE certification \cite{kostolani2019effective}), but can say little about the safety assurance of the factory operation as a whole as they do not have the necessary knowledge. It is the responsibility therefore of the system integrator to assess whether the overall safety of the system can be demonstrated by the integration of the components. In particular, the integrator must identify, through safety analysis, the hazards posed by factory operation and determine how the components may contribute to hazards (this could for example be done through considering deviations on the functionality of the components and their interactions).

As a result of this delegation of safety responsibility to different stakeholders, we determined that a modular safety case approach should be adopted \cite{kelly2001concepts}. In a modular safety case, the overall safety case for the dynamic factory is split into separate modules of argument and evidence relating to different system components. 

\subsection{Safety Contracts}

The safety case modules can be linked together using safety contracts in order to create a coherent case for the overall system. Such a modular approach requires that the safety contract for each component in the factory be determined and explicitly defined. This contract defines the set of properties that the component is able to assure and a definition of potential failure behaviour of the component. In order to be usable as part of the integrated safety case for the factory, we propose that each of the identified properties should be defined with the following assume-guarantee reasoning form:

\begin{quote}
\emph{if \{condition\} then \{component\} shall provide \{property\} with confidence of \{confidence\}}
\end{quote}

The \emph{condition} and \emph{property} represent, respectively, the assumptions and guarantees of an assume-guarantee contract \cite{sangiovanni12}. In particular the \emph{condition} and \emph{confidence} of this assume-guarantee contract specification is crucial to our approach; for any component there exist limitations on the circumstances under which its behaviour is guaranteed and these must be clearly understood and specified. Understanding and expressing these limitations is particularly important in systems that make use of COTS components, where the confidence may be quite low. The conditions that affect a guarantee could be diverse, depending on the nature of the guarantee and the component. Conditions may relate to the state of the operating environmental, such as the effects of lighting conditions on sensor performance. They may be conditions on other system components, such as the availability of electrical power. Conditions may also include operational constraints on the way a system or component is installed or operated. 

As an example, an assurance contract for a pressure sensor may include:
\begin{quote}
\emph{If temperature is greater than -20$^{\circ}$C then pressure sensor shall provide air pressure value with accuracy of 0.1\% with confidence of 99\%.}
\end{quote}

The construction of such a safety contract requires evidence for the accuracy of the pressure sensor, for example evidence from testing of the sensors and in-service data for the sensor performance in operation. The more evidence that is available, the higher the confidence in that performance will be. This evidence would be included as part of the safety case module for the pressure sensor. The safety contract also reflects the knowledge that the sensor providers have that the sensors do not perform as well at very low temperatures. This places the limitation condition on temperature defined in the contract. Using the safety contract, the integrator will be able to assess whether a pressure sensor meeting this safety contract will be sufficient to meet the safety requirements identified for the overall system.

In this paper we analyse the safety requirements for an experimental scenario involving the use of geofencing technology in the operation of an AGV in a factory. The results highlight the need for the development of a modular safety case and help us determine the required safety contracts for key elements of the system.

\section{Factory AGV Geofencing Scenario}\label{sec:scenario}

Our scenario considers a highly automated and flexible factory that includes AGVs that move inventory and equipment around a factory. Due to the dynamic reconfiguration that may occur in the factory, the AGVs cannot simply follow fixed routes around the factory, but must instead be capable of planning their own paths through the factory in order to fulfill their task. The AGVs must determine the best path to safely perform a task without colliding with objects in the factory environment (such as equipment, machinery, humans or other AGVs), or entering restricted areas (such as areas of high human occupancy). Achieving this requires sensing, perception and path planning technology on the AGVs. There are three techniques that combine to achieve this:

1. Localisation: (see Hodge \cite{hodge2022} and Zafari et al. \cite{zafari2019survey} for surveys of techniques). This harnesses the ubiquitous connectivity of Internet of Things (IoT) to localize and position IoT devices (tags). This is suitable for dynamic factories to set up safe areas with restricted access (geofencing), e.g., areas under maintenance, where humans operate or where the AGV would be unsafe, e.g. a spillage. It gives a global overview of the AGV and its navigation environment but cannot locate obstacles or other collision points (local view).

2. Local motion planning: Using object detection/recognition \cite{Hodge2020}. This requires AGV-mounted sensors such as IR proximity sensors, LIDAR, or cameras. This navigation is suitable for detecting dense objects such as relocatable machinery, boxes, other AGVs, vehicles. It generates a detailed and high fidelity map of the AGV's immediate surroundings. It provides a fine-grained local view but not a global overview of the navigation so the AGV cannot avoid danger zones or areas where humans are working. 

3. Global path planning: Generally uses a map which generates a fixed overview and may require prior knowledge of the layout. It is ideal for large and immovable objects. The map can be a) pre-defined and fixed such as a floorplan or CAD drawing; or b) constructed by the AGV itself using SLAM ``to incrementally build a map of this environment while simultaneously using this map to compute absolute vehicle location'' \cite{dissanayake2001solution}. This is more dynamic but is slow and compute-intensive. In \cite{Hodge2020} we developed a mapless navigation algorithm that uses only data from on-board sensors (local data) to navigate; and analysed its safety.

A combined navigation approach would overcome the weaknesses in each technique and achieve both a detailed local view and an overall view of the navigation. It helps ensure no single point of failure.  In \cite{Hodge2020} we developed and assured global path planning. As a further step to achieving a combined approach, we focus on ensuring the safety of  1) localisation  using fog/edge geofencing to create an overall view, to provide protection for areas of the factory where humans work and stop AGV encroachment into danger zones. Geofenced areas can be created and removed when needed (e.g. during emergency repairs or where humans are working) and are more flexible than maps as comparing an AGV's location against a pre-defined map cannot cope with dynamic layout changes.  The geofencing can be used to provide an emergency stop capability for the AGV. 

\begin{figure}
\centering
\includegraphics[width=6.5cm]{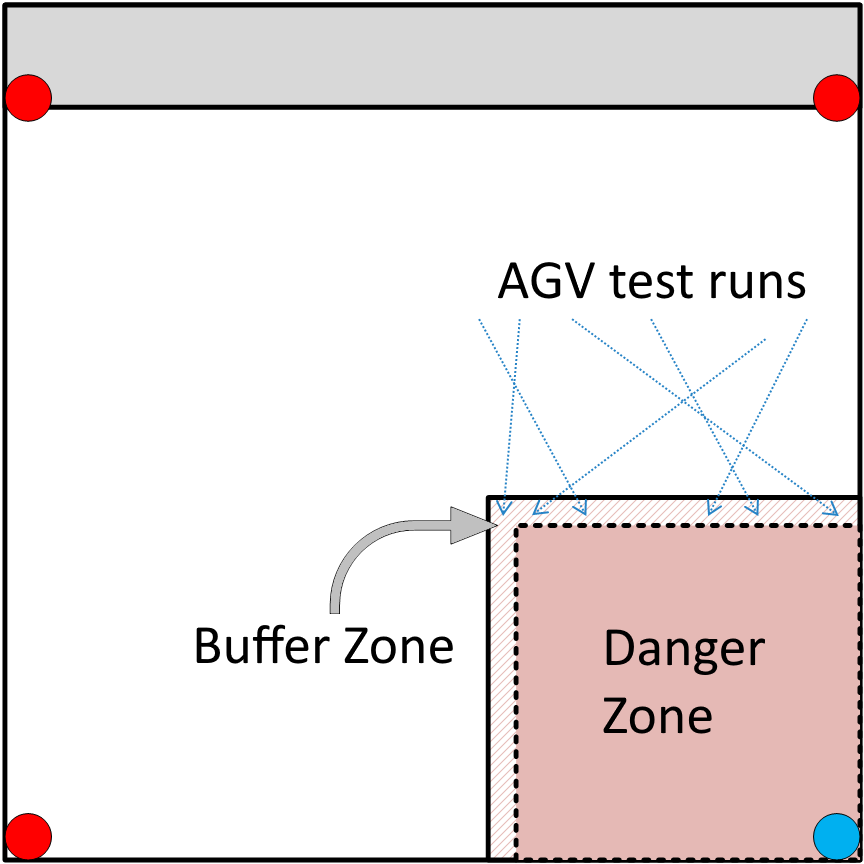}
\caption{Our scenario is 8m x 8m x 2.5m high with four geofencing anchors in a square configuration (master anchor bottom right). The geofenced zone is the danger zone (bottom right) plus 30cm buffer zone where the AGV stops.\label{fig:robotDanger}}
\vspace*{-1.5mm}
\end{figure}

The core requirements of the geofencing are: 
\begin{enumerate}
\item The AGV should STOP before entering the danger zone.
\item If the AGV enters the danger zone, then it should be immediately stopped.
\item The AGV should navigate away from the buffer zone.
\item Accurate geolocation of tags placed on the AGV.
\item The system should detect possible failures of hardware, software and comms and ensure ``Safe Failure''. If this affects an AGV then that AGV should stop. If it affects all AGVs (network outage) then they should all stop.
\end{enumerate}
Criteria 3 and 5 are not considered in our initial work here as they relate to the AGV's path-planning abilities rather than the geofencing. Fig. \ref{fig:robotDanger} shows a schematic of our UWB geofenced laboratory where AGVs must not encroach into the danger zone and must obey safety requirements.

\subsection{Safety Requirements}\label{sec:safetyReq}

By considering the hazards associated with the scenario described above the following high-level safety requirements can be identified for the AGV system:

\begin{itemize}
\item \textbf{SR1} – AGV shall not plan a path that results in collision with an object in the factory
\item \textbf{SR2} – AGV shall stop before entering an unsafe zone (danger zone in Fig. \ref{fig:robotDanger}) or colliding with an object
\end{itemize}

SR1 says that when planning a path in order to achieve a task within the factory (such as moving parts from one location to another) the AGV should ensure that its path does not result in a collision. This can be achieved using information from sensors and object detection to locate objects \cite{Hodge2020}. SR2 says that even if an unsafe path is planned, the AGV should not enter a danger zone or collide with an object. This requirement can be considered as an ``emergency stop'' or ''collision avoidance`` facility, which provides extra safety assurance. 

This second safety requirement is the focus in this paper. We focus on an ``emergency stop'' with geofencing. This allows the AGV to avoid dangerous areas that a sensor-based collision avoidance system would not be able to `see'. Sensor-based collision avoidance is covered extensively in the literature, e.g., \cite{guiochet2017safety}. We consider how geofencing implemented using a real-time indoor location system (RTLS) can provide assurance against SR2. 
We analyse how modular safety criteria can be developed. This entails determining safety contracts for each of the components of the RTLS, and generating evidence to support the satisfaction of the safety contracts. This involves the following steps:

\begin{itemize}
\item identifying the properties that can be guaranteed for the components
\item generating evidence to support these properties
\item understanding the conditions under which the properties hold
\item identifying failure conditions for the components
\end{itemize}

\subsection{Geofencing Implementation}\label{sec:geofenceImpl}
One emerging RF technology for accurate localisation is ultra-wideband (UWB) ranging technology which uses tags and anchors for geolocation. We provide a literature review of UWB geofencing in section \ref{sec:related}. UWB devices transmit ultra short-pulses over a large bandwidth ($>$500MHz), with frequency range between 3.1 and 10.6GHz, and a very
low duty cycle. This reduces power consumption \cite{zafari2019survey}. UWB is robust to interference from other signals \cite{geng2005multipath}. 

Our experimental setting is based on Sewio's Indoor UWB RTLS Kit \cite{sewio} for AGV geofencing, analysed against our safety requirements, determining what factors affect its ability to meet those requirements. 
 It is widely used in industry, can support up to 5000 tags and has a reported positioning accuracy of +-30cm (98\% of positions have an error $<$30 cm). Hulka et al. \cite{hulka2020accuracy} found the Sewio RTLS UWB tracking system accurate and reliable for tracking basketball players. Contigiani et al. \cite{contigiani2018} compared the Sewio and Openrtls UWB tracking systems and found similar positioning accuracy but Sewio had superior battery life due to its energy saving. However, they found positioning accuracy degraded when used in an occluded environment due to NLOS errors. Hence, we evaluate NLOS tracking in section \ref{sec:eval}.

Our RTLS geofencing system has five modules \textbf{1) Tags} mounted on the AGVs and powered by rechargeable Li-Ion batteries. The tags transmit data packets periodically over Decawave UWB radio to data receivers (anchors). The tags do not communicate peer-to-peer. \textbf{2) Anchors} mounted on brackets in fixed positions (see Fig. \ref{fig:robotDanger}) which are reference devices with a known position. One anchor is designated the ``main''. The system uses triangulation and the precise measurement of time difference between a
signal arriving from a tag (blink) to a set of anchors to calculate the tag's exact location. To ensure high accuracy, the anchors need to be
synchronized very precisely which is done periodically from the master. \textbf{3) Location Engine} software that collects the tag and anchor data and calculates the absolute tag positions. \textbf{4) Safety Controller}. We developed this in Java using Sewio's Java API \cite{sewioAPI} to calculate the relative distance between tags and geofenced zones. If a tag encroaches into a zone, the controller generates an ``alert'' and sends a ``STOP'' signal (see section \ref{sec:postFilter}). Although we only analyse two AGVs here, having the controller on a laptop (or laptops) connected to the main anchor via wi-fi router allows us to generate an overall picture of a factory. This controls the AGVs and factory safety and can also incorporate global maps and map-based control. If we had a safety controller on each AGV then we would need to update all AGV controllers every time there was a factory layout change and would have difficulty controlling the AGVs as a group. \textbf{5) User Interface},  currently a simple command line interface that runs in conjunction with the safety controller and is designed to display tag locations and geofenced zone alerts.



\section{Evaluating safety contracts for UWB Geofencing}\label{sec:evaluating}

From SR2 we know that our system described above must provide a stop command to the AGV in sufficient time for it to stop before entering the danger zone. In this section we describe experiments conducted to determine the nature of the safety contract for our system by identifying the safety guarantees that it is possible to make for the system, as well as the conditions under which those guarantees hold.

\subsection{Experiment Design}
The purpose of our experiment is to determine if properties can be guaranteed by the fog/edge UWB system in the defined scenario and to identify what factors impact on the ability of the UWB system to provide these guarantees. This will allow us to develop the safety contract for the UWB geofencing system. We will also use to results of the experiment to understand the dependencies on the UWB system components and their safety contracts.

\subsubsection{UWB Indoor Geofencing System}
As we are making our initial investigation of safety and developing safety assurance, the entire real-time location system was implemented and evaluated in the University of York Robotics Lab~\cite{yrl},  a purpose-built research facility. 

Sewio provided detailed instructions on how to install the system correctly. We ensured we followed these, placing the anchors in a square configuration all around 2.5m off the ground to ensure consistent and reliable coverage, using Sewio's software and charts to optimise the signal strengths and synchronisation stability between anchors. We ensured that the tags were at least 15cm above the floor at all times (by placing them on a cardboard box on top of our AGV) to ensure height compliance and to keep them away from metal objects on the AGV which can cause interference. As stated earlier, Sewio tags have a claimed positioning accuracy of +-30cm so we set the virtual stop line 30cm back from our physical stop line giving a buffer zone where the AGV should stop (shown in Fig. \ref{fig:robotDanger}) with 98\% confidence. 

In section \ref{sec:eval}, we analyse this setup for accuracy and latency. Sedlacek et al. \cite{Sedlacek2016} found that Sewio tags had a median positioning error of 39.6cm at various locations including those with diminished anchor coverage. The errors varied with location. A tag placed in the centre of the coverage area had median positioning error of 23.0cm but one near the corner with line-of-sight of fewer anchors had median error of 73.8cm \cite{Sedlacek2016}. In our future work, we will analyse varying the size of the buffer zone allocated to each AGV according to that AGV's specifications (velocity, size, tag transmission rate etc.). Here, we focus on developing safety assurances so mandate that all AGVs have identical setup so we can investigate what specifications ensure the AGV stops in time (and what specifications cannot ensure the AGV stops) and thus what guarantees we can make.

\subsubsection{Safety Controller}\label{sec:postFilter} We need to detect encroachments using the AGV tag's location and issue a safety alert to the AGV so it stops. For this, we used a MS Windows laptop  connected to the Sewio wi-fi network. It runs the Sewio VirtualBox VM image \cite{sewio} and provides a bridge between:\\ (tags+anchors) $\leftrightarrow$ (Cisco router) $\leftrightarrow$ (Java API in the VM).

We setup one geofenced zone (as shown in Fig. \ref{fig:robotDanger}) via Sewio's software. Our Java Safety Controller (section \ref{sec:geofenceImpl}) analyses tag and geofence information extracted using Sewio's API \cite{sewioAPI}. It provides command-line alerts to the user and issues  ``Stop'' commands to the AGV if it encroaches. The communication from laptop to the AGV uses plink~\cite{plink} - a command-line tool from MS Windows (laptop) to Linux (AGV) that executes remote commands on the AGV via a non-interactive \emph{ssh} session. This ensures minimal processing overhead and so we can pinpoint delays more easily.  

\subsubsection{ Control and Communication with AGV and Machine}
Our evaluation used two purpose-built AGVs based upon the York Robotics Kit hardware~\cite{yrlOcto}; \textbf{R1} controlled by a Raspberry Pi 3B+ and \textbf{R2} controlled by a Raspberry Pi 3A. They have different speeds of travel. R1 defaults to 0.093 m/s and R2 to 0.277 m/s, i.e., R2 is 3x faster than R1. This provides the opportunity for evaluating different speeds and its effects. Using COTS AGVs ensures we have total control over the AGVs (full software access). COTS AGVs are also more accessible. Many commercial AGVs are out of the price range of all but high-end factories. We aim to demonstrate that we can assure safety across the range from COTS to high-end commercial AGVs. We start here by analysing and ensuring the safety of COTS AGVs. 
The AGVs run York Robotics Lab Python control software to control the motor speeds \cite{yrlPy}. The Java Safety Controller on the laptop uses plink to call the Python running under Linux on the AGV.  Each robot travels on a variety of (randomly chosen) trajectories towards the protected zone to thoroughly analyse the stopping conditions as shown by the blue dotted arrows in Fig. \ref{fig:robotDanger}.






\subsection{Experimental Results}\label{sec:eval}

To analyse the safety guarantees of the geofencing, we set the tags to default settings: transmission every 100ms with medium signal strength to ensure battery life without compromising  signal quality. We ran 2 evaluations.




\subsubsection{Non-Line-of-Sight (NLOS) test}\label{sec:NLOS} 
We investigated NLOS detection by running R1  with a tag mounted on top into a tunnel constructed from three (thick-walled) cardboard boxes. The tunnel was 39cm wide, 31cm high, 43 cm deep and placed straddling the virtual stop line.  While inside the tunnel, the tag functioned normally while the AGV travelled through the tunnel and the AGV stopped in the same range (within the box plots in Fig. \ref{fig:stopDists}) as when the AGV was not in the tunnel.

\subsubsection{AGV Configurations test}\label{sec:configTest}

\begin{table}
\begin{center}
\caption{Table listing the AGV test configurations and whether the AGV stopped safely outside the danger zone. Tag transmits every $n$ milliseconds (Tag Period). $v$ is max speed of R2, $v/3$ is max speed of R1 (and R2 running at 1/3 speed). Server indicates if the Python data server was running. Rand indicates whether a random offset is added to the gaps between tag transmissions.}
\label{table:1}\begin{tabular}{ c|c|c|c|c|c|c} 
 \hline
Test ID & AGV & Tag Period & Speed & Server & Rand & Stopped?\\ 
 \hline
 T1 & R1 & 100 & v/3 & N &  Y  & yes \\ 
 T2 & R2 & 100 & v   & Y &  Y  & no\\ 
 T3 & R2 & 100 & v/3 & Y &  Y  & no\\ 
 T4 & R2 & 100 & v   & N &  Y  & no\\ 
 T5 & R2 & 100 & v/3 & N &  Y  & yes\\ 
 T6 & R2 & 200 & v   & N &  Y  & no \\ 
 T7 & R2 & 200 & v/3 & N &  Y  & no\\ 
 T8 & R2 & 200 & v   & N &  N  & no\\ 
 T9 & R2 & 200 & v/3 & N &  N  & no (once)\\ 
 \hline
\end{tabular}
\end{center}
\end{table}

\begin{figure}[h!]
\centering
\includegraphics[width=9cm]{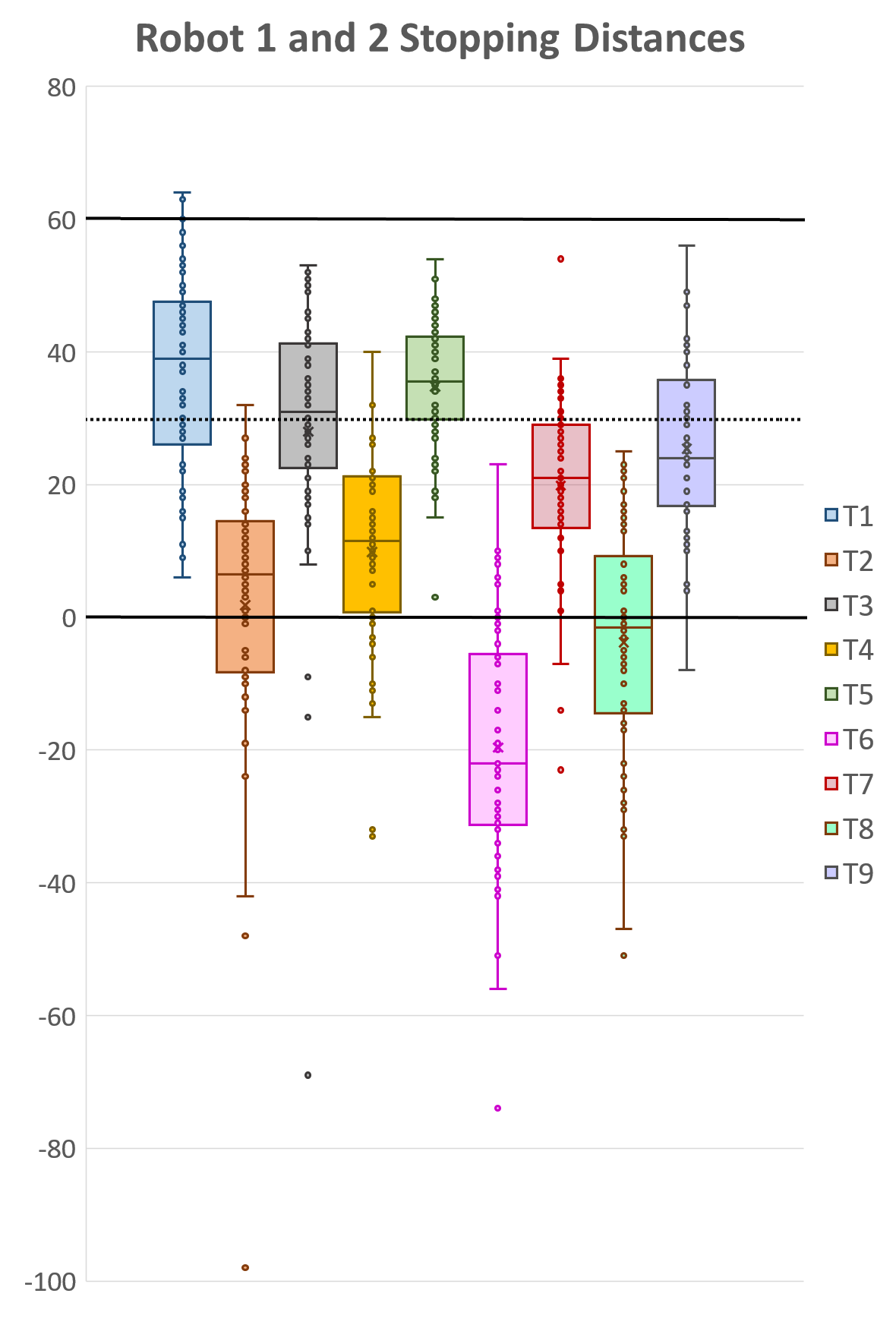}
\caption{Box and Whisker Plots of the stopping distances (measured perpendicularly from the stop line with a tape measure) for test configurations (T1-T9) over 50 runs per configuration. 0cm is the danger zone. +30cm is the virtual stop line to give a stopping buffer. 60cm is the outer edge of the buffer. AGVs should stop before 0cm line on all runs.\label{fig:stopDists}}
\vspace*{-1.5mm}
\end{figure}

We drove the AGVs at the danger zone at a range of angles (see Fig. \ref{fig:robotDanger}). The AGV stops when the system detects an encroachment. We measured the distance from the front of the tag to the front of the physical stop line (black dashed line shown in Fig. \ref{fig:robotDanger}), repeating this process 50 times (over a number of days) for the nine AGV configurations given in Table \ref{table:1}. By varying the configurations, we evaluate the following criteria:
\begin{itemize}
    \item AGV speed where $v$ is max speed of R2, $v/3$ is max speed of R1 (and is R2 running at 1/3 speed).
    \item processor / RAM. R1's Raspberry Pi has a Cortex-A53 (ARMv8) 64-bit @1.4GHz with
1GB LPDDR2 SDRAM while R2 has the same CPU with
512MB LPDDR2 SDRAM. Both run Linux (and Python).
    \item effect of running processes - run a Python data server on the Raspberry Pi that hogs the CPU periodically.
    \item tag transmission period (in ms).
    \item regularity of tag transmission.
\end{itemize}
We list if the AGV stops before the danger zone for each configuration in Table \ref{table:1} and create box and whisker plots for the AGVs' stopping distances in Fig. \ref{fig:stopDists} (distance between tag and boundary of danger zone which should be $>$0cm). The box and whisker plots show the median, inter-quartile range (box), upper and lower extremes (whiskers) and outlier points (dots). To meet the safety guarantee each box, whisker and dots should be above the y=0 line in their entirety. 

\subsection{Using Results to Define Safety Contracts for Geofencing System}

From experiments we can identify the properties that can be guaranteeed by the geofencing system along with the conditions under which those properties can be guaranteed. In our case study, with relation to safety requirement SR2 (see section \ref{sec:safetyReq}), the property of concern for the safety contract is the \textbf{stopping distance}. This requires the following guarantee: \textit{The AGV will stop no more than 30 cm beyond the defined virtual stop line (giving a buffer zone).} We need to analyse the results of our case study to determine the factors that guarantee this. 


From the results in Table \ref{table:1} and Fig. \ref{fig:stopDists} we find  that only T1 and T5 guaranteed this result over 50 runs. Analysing further, with 99\% confidence the population mean stopping distance (assuming a population of 1M runs) is between 31.3 and 42.2cm based on 50 samples for T1 (36.74 $\pm$ 5.46cm before the danger zone), and with 99\% confidence the population mean stopping distance is between 31.1 and 38.4cm based on 50 samples for T5 (34.76 $\pm$ 3.68cm before the danger zone). T1 uses AGV R1; and  T5 uses R2 with the same configuration as R1 had for T1. This provides a preliminary analysis of safety guarantees for geolocation of AGVs. 
If we want a confidence level of $99\% \pm 1$ ($>$98\%) then we need to analyse each configuration over 16369 runs to provide true safety guarantees for a safety-critical task or environment.

Even over 50 runs, we can identify factors of the AGVs that would fail to guarantee safety. T9 encroached once into the danger zone (the lower extreme of the box and whisker is below the stop line in Fig. \ref{fig:stopDists}) but this fails the safety guarantee nonetheless. With 99\% confidence the population mean of T9 is between 20.6 and 30.1cm, based on 50 samples - this fails SR2. All other test configurations encroach multiple times. The variables that were seen to affect these results were: AGV velocities, running processes on AGVs, tag transmission periods (gaps) and tag transmission regularity. In particular, having a server running on the AGV caused extreme outlier values in stopping distances and large encroachments as did a combination of faster AGV speed and slower tag transmission rates. Switching off random offset on the tag transmission gap also caused large encroachments on the faster AGV. NLOS operation did not affect the AGV stopping distance in our analyses though other authors found that NLOS did adversely affect geolocation \cite{contigiani2018,zandian2016performance}. These observations allow us to identify the conditions for safety assurance and plan our future work such as dynamic buffer zones (see section \ref{sec:future}) related to AGV velocity.

Based on our case study we can therefore identify the following conditions upon which the guarantee relies:

\begin{itemize}
\item AGV speed does not exceed 0.093 m/s
\item Control command software process has schedule priority on AGV Raspberry Pi
\item Tag transmission gap is at least 100 milliseconds
\item There is a random offset between tag transmissions

\end{itemize}

\begin{figure}[h]
\centering
\includegraphics[height=8.5cm]{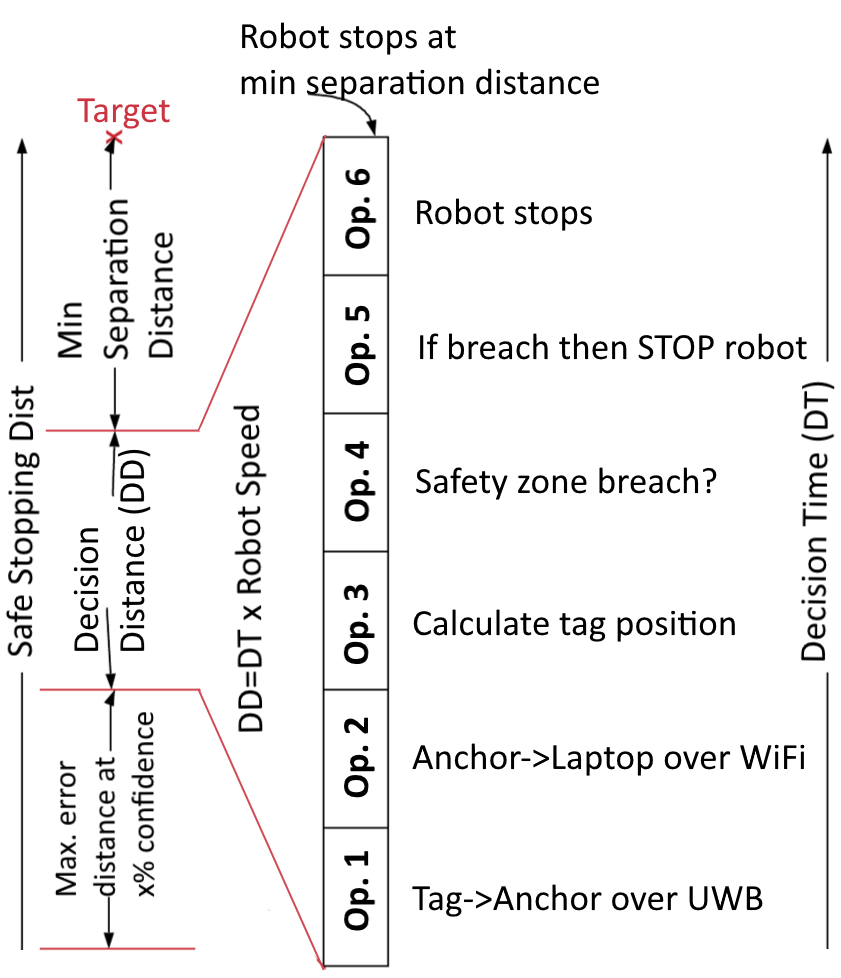}
\caption{Stages for geofencing control. \label{fig:stages}}
\end{figure}

In fact, by considering geofencing as an end-to-end service, we can see that it consists of a number of stages. Fig. \ref{fig:stages} illustrates the stages required for AGV control using geofencing. A safety contract could be specified for any of these stages, defining the individual properties that can be guaranteed, and the conditions under which they hold. This illustrates why a modular safety approach is required; properties cannot be guaranteed for geofencing unless properties of the individual components (wifi, comms, tags, anchors, processors) can also be guaranteed. For example the tags must guarantee with sufficient confidence that 100 millisecond transmission gaps can be provided. The conditions that are required for this must be specified as part of the contract for the tag. These conditions in turn must be demonstrated to hold. Adopting a modular safety approach also supports changes to the system and the configuration of the factory. The explicit safety contract for each system element defines the bounds within which that element must perform. When changes are made to any individual element, as long as the element can still be shown to satisfy the safety contract, we know that the overall safety requirement can still be met without the need to re-evaluate the whole system. This would allow, for example, different tags with different performance to be used, so long as the guarantees specified by the contract are still met.

\section{Related Geolocation Work}\label{sec:related}

Other authors have analysed various aspects of UWB geolocation. Zandian and Witkowski  use similar performance analyses to ours but they built their own UWB system rather than using a commercially available system \cite{zandian2016performance}. Segura et al. \cite{segura2010experimental} also built a real-time AGV navigation system for indoor
environments using COTS
components. We want to verify industry-standard UWB kit and can incorporate industry standard AGVs as we develop our safety assurance framework and safety-assured system. 
For the initial analyses here, it is much easier to control, assess and verify our COTS AGVs. Ruiz and Granja \cite{ruiz2017comparing} compared three commercial UWB systems for accuracy in  an experimental evaluation and found a range of performances with Decawave outperforming BeSpoon which in turn outperforms UbiSense.  The Sewio kit we analysed uses Decawave hardware. 

Martinkovi{\v{c}} et al. \cite{martinkovivc2019use} investigate a real-time location system for hybrid assembly environments. They analyse how UWB tracking can be incorporated in manufacturing where humans and AGVs cooperate closely.  Similarly, Park et al. \cite{park2016bim} combine UWB tracking with dead reckoning and a path planner for safe AGV navigation in indoor construction sites.
Ridolfi et al. \cite{ridolfi2018analysis} look at the scalability of indoor UWB tracking systems and how to maximise the number of tags that can be tracked. Krishnan et al. \cite{krishnan2007uwb} focus on optimising the network setup for best localisation accuracy. Ieni \cite{ieni2018realization} analyses the performance of indoor UWB tracking by evaluating different localization algorithms. These systems provide practical localisation results for analysis but do not provide safety-assessments or safety guarantees. 

{\v{C}}ernohorsk{\`y}
et al. \cite{cernohorsky2018real} investigate UWB tracking in confined spaces (e.g., corridors) and provide suggestions for overcoming issues with interference and accuracy. Zandian et al. \cite{zandian2016performance} compare the performance of LOS vs NLOS UWB tracking. They find a slight degradation in positioning accuracy with NLOS compared to LOS geolocation. These authors findings will need to be considered in our future work on developing safety contracts for dynamic factories.

Other uses of UWB tracking include tracking sports players such as basketball players \cite{hulka2020accuracy} for player movement analysis, tracking shoppers inside retail stores to analyse their behaviour (movements) \cite{contigiani2018} \cite{gabellini2019large}, tracking players in large-scale augmented reality (AR) systems  \cite{cirulis2019}, incorporating tracking tags in AR headsets to geolocate users \cite{cyrus2019hololens} or tracking indoor drones in warehouses \cite{shule2020uwb,macoir2019uwb,tiemann2015design}.

\section{Conclusion and Future Work}\label{sec:future}
This paper introduces a case study for evaluating the safety assurance of AGVs in dynamic factories. We analyse the accuracy of geolocating AGVs using fog/edge computing and whether an AGV can stop outside a virtual danger zone set up using a commercial UWB tracking system to protect the area from encroachment. In particular we investigated safety requirement SR2 from section \ref{sec:safetyReq}, \textit{AGV shall stop outside the danger zone and before colliding with an object}. We found that the safety contract conditions rely on guarantees made by different components of the UWB system (tags, anchors, AGV speed and on-board processors which affect AGV response time) and fog/edge (wi-fi) capabilities - these components will all require safety contracts. Additionally, AGV geofencing is an end-to-end service that requires a number of stages. Each stage could have a separate contract defining the properties that can be guaranteed and the conditions under which they hold, supporting a modular approach to safety assurance. 

This study is a proof-of-concept. Further work will analyse the geofencing and guarantees over more AGV runs and extend the analyses to navigating the AGV away from the danger zone and detecting failures of software and comms. We will perform load testing to investigate how many AGVs can be safely controlled in conjunction with how many safety controller laptops are necessary for a given number of AGVs. We can further investigate the fog/edge network and its effects on stopping distances. How do we certify the network? What limits do we need to impose? How can we overcome network errors and still ensure safety?

Future work will develop a set of contract templates and establish a set of heuristics for guaranteeing that the contracts can be met (e.g., min/max AGV speed, tag transmission rate). We will evaluate a range of AGVs (COTS to commercial AGVs) to establish these parameters and also investigate using a variable-depth buffer zone for different AGVs. The radius of the buffer will vary to reflect how the guaranteed stopping distance changes with different AGVs and according to the current state of an AGV (its speed, tag transmission gap, wi-fi network throughput, network delay etc.)

Further ahead, we can combine the UWB geolocation with global path planning using maps and algorithms such as A* and mapless path planning using sensor data only~\cite{Hodge2020}.






\section*{Acknowledgement}
The work is funded by the Swedish Foundation for Strategic Research
under the project “Future factories in the cloud (FiC)” and the Assuring Autonomy International Programme (www.york.ac.uk/assuring-autonomy).

\bibliographystyle{ieeetr}
\bibliography{biblio}

\end{document}